\documentclass[10pt,twocolumn,letterpaper]{article}

\usepackage[english]{babel}
\usepackage{algorithm}
\usepackage{algorithmic}
\usepackage{abstract}
\usepackage{graphicx}
\usepackage{color}
\usepackage{transparent}
\usepackage{verbatim}
\usepackage{amsmath}
\usepackage{capt-of,etoolbox}
\usepackage[caption=false]{subfig}
\usepackage{paralist}
\usepackage{amsthm}
\usepackage{enumitem}
\usepackage{amsfonts}
\usepackage{iccv}
\usepackage{times}
\usepackage{epsfig}
\usepackage{amssymb}
\usepackage[autostyle, english = american]{csquotes}

\usepackage{array}

\MakeOuterQuote{"}

\iccvfinalcopy

\theoremstyle{definition}

\begin{document}

\title{Detecting Moving Regions in CrowdCam Images}
\author{Adi Dafni\\
  Tel-Aviv university, Israel\\
  adidafni@gmail.com\\	
\and Yael Moses \\
  The Interdisciplinary Center, Israel\\
  yael@idc.ac.il
	\and Shai Avidan\\
  Tel-Aviv university, Israel\\
   avidan@eng.tau.ac.il}

\renewcommand{\tabcolsep}{1pt}
\makeatletter
\let\@oldmaketitle\@maketitle
\renewcommand{\@maketitle}{\@oldmaketitle%
\begin{center}\begin{tabular}{c c c c } 
	\includegraphics[width=0.2\linewidth]{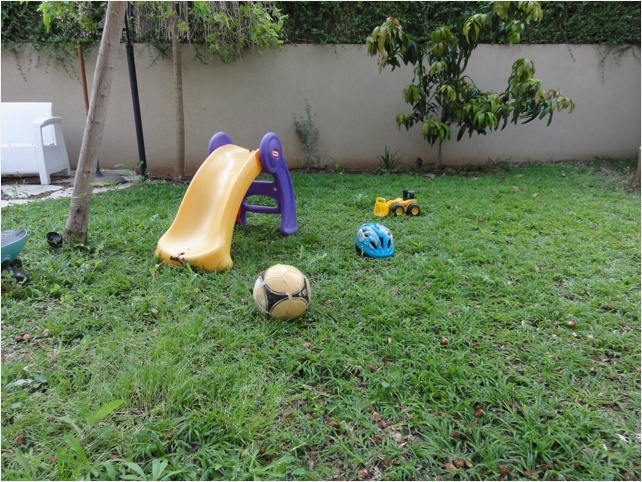}&
		\includegraphics[width=0.2\linewidth]{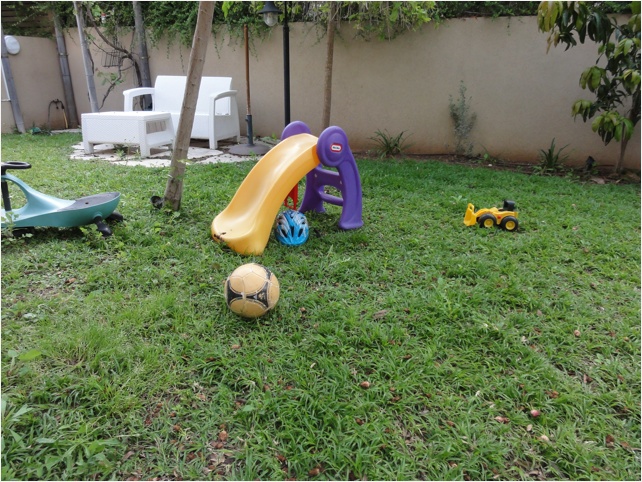}&
		\includegraphics[width=0.2\linewidth]{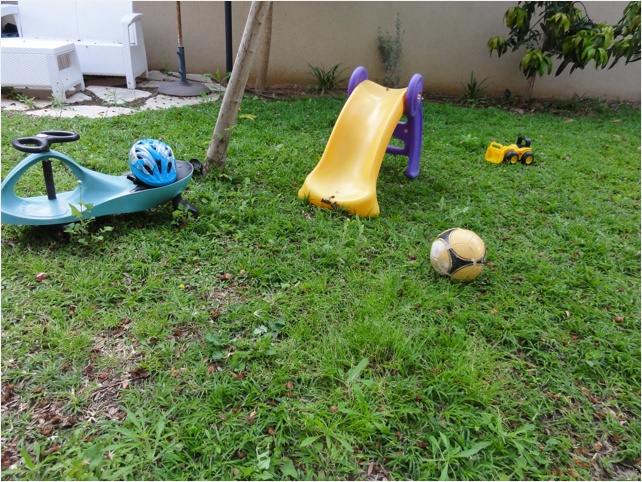}&
		\includegraphics[width=0.2\linewidth]{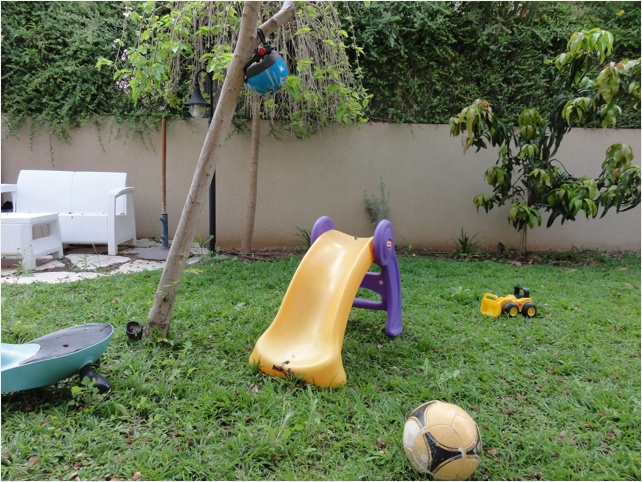}\\
\end{tabular}\end{center}
Figure~1: A set of images captured at different times by different cameras. Which object moved? (See Fig~\ref{fig:results} for our results.) 
	\bigskip} 
\makeatother
\maketitle

\begin{abstract}
We address the novel problem of detecting dynamic regions in CrowdCam images -- a set of still images captured by a group of people.
These regions capture the most interesting parts of the scene, and detecting them plays an important role in the analysis of visual data. 
Our method is based on the observation that matching static points must satisfy the epipolar geometry constraints, but computing exact matches is challenging. Instead, we compute the probability that a pixel has a match, not necessarily the correct one, along the corresponding epipolar line. The complement of this probability is not  necessarily the probability of a dynamic point because of occlusions, noise, and matching errors. Therefore, information from all pairs of images is aggregated to obtain a high quality dynamic probability map, per image. 
Experiments on challenging datasets demonstrate the effectiveness of the algorithm on a broad range of settings; no prior knowledge about the scene, the camera characteristics or the camera locations is required.
\end{abstract}

\section{Introduction}

CrowdCam images are images captured by a crowd of people. These images usually capture some interesting dynamic event, and the dynamic objects are often where the attention should be drawn. It is therefore useful to ask whether the dynamic regions of a scene from CrowdCam images can be detected. A method for detecting these regions  can be used to propose image windows that are likely to contain an object of interest. Other computer vision applications that can benefit from such a method include change detection, moving object segmentation, and action recognition.

In this paper, we address the novel problem of detecting the dynamic regions in a scene from CrowdCam images. As these images may be taken with a wide baseline in space and time, significant new challenges arise, such as  distinguishing an object that moved from one whose appearance changed due to changes in the camera's viewpoint or occlusions (as demonstrated in Fig.~1).

\begin {figure*}[t]
\centerline{
	\begin{tabular}{c c c}
		\includegraphics[width=0.31\linewidth]{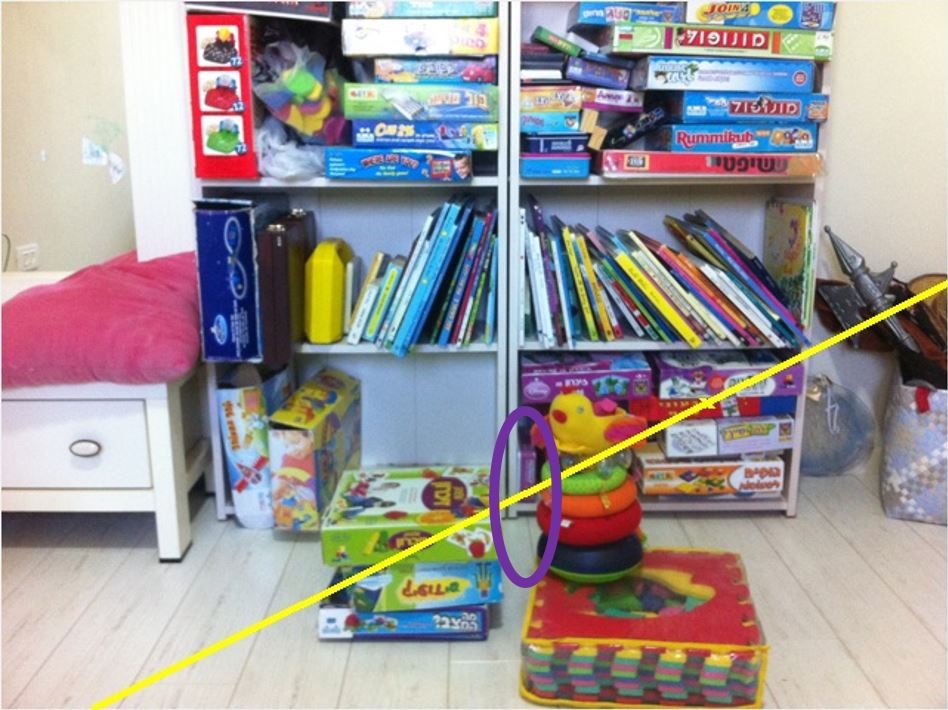}&
		\includegraphics[width=0.31\linewidth]{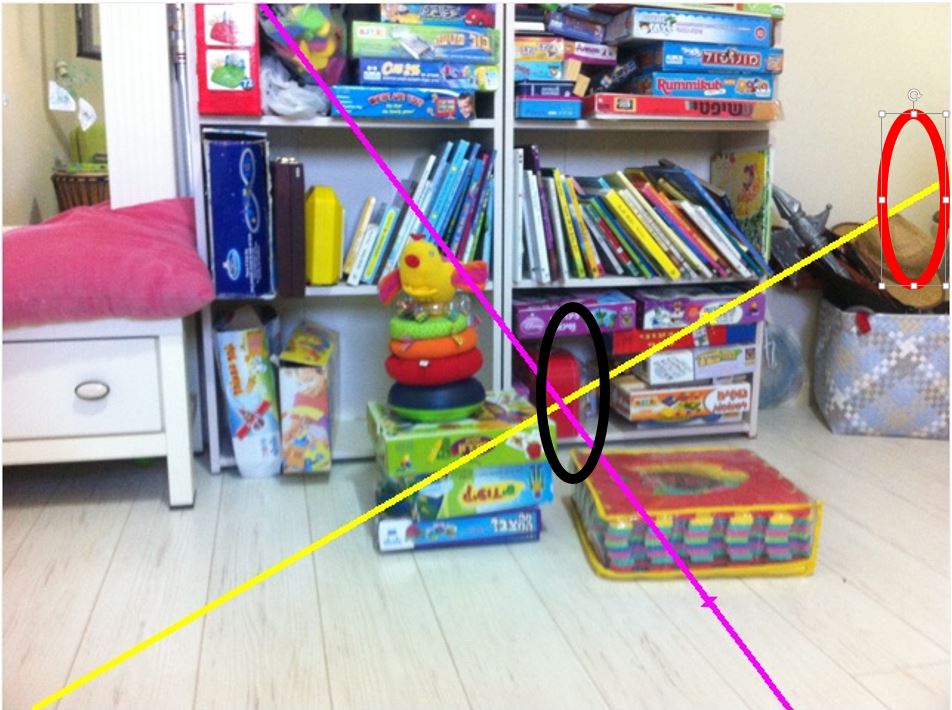}&
	  \includegraphics[width=0.31\linewidth]{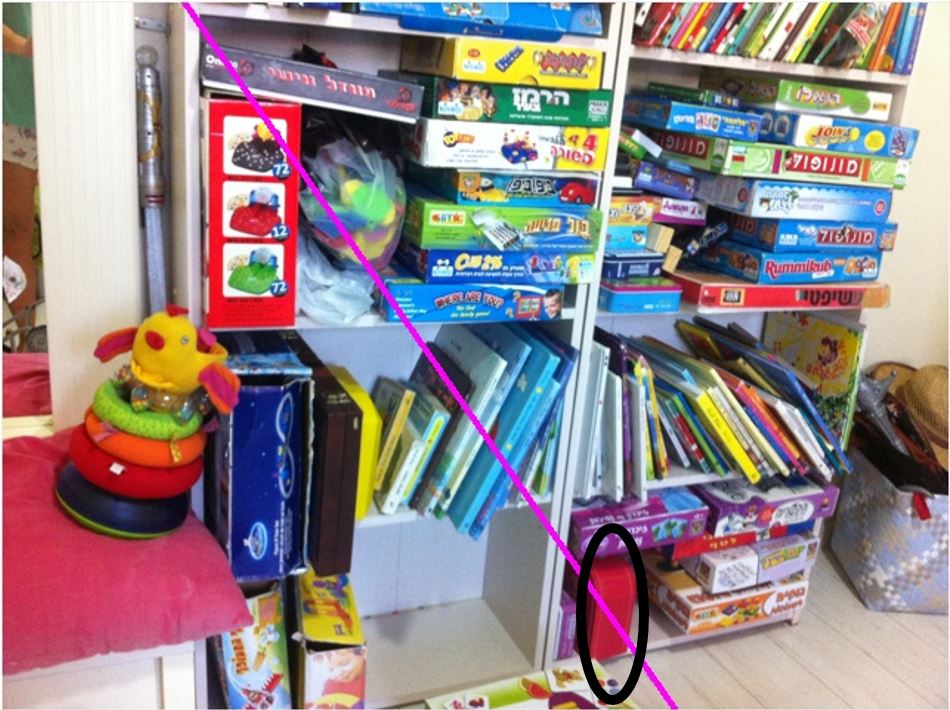}\\
			(a) & (b) & (c) \\
	\end{tabular}}
	\caption{Types of occlusions in a wide baseline image pair with a moving object: Consider images (a) and (b) -- corresponding epipolar lines are marked in yellow, ellipses indicate locations that are occluded by (i) an out of field of view (red), (ii) different viewpoints (purple), (iii) a moving object such as the  chicken toy (black). When image (c) is considered as well (corresponding epipolar lines between (b) and (c) are marked in magenta), the red suitcase correspondence  occluded in (a) by the moving object is revealed. }
\label{fig:occlusions}
\end {figure*}

CrowdCam images violate the assumptions made by existing methods for detecting moving regions. Background subtraction methods assume the camera is static, or at least that the images can be properly aligned. 
Motion segmentation algorithms usually work on video, which has a high temporal frame rate and small baseline between successive frames. In CrowdCam images, on the other hand, the images are few and far between, thus they cannot necessarily be aligned, making it impossible to preserve  the spatial-temporal continuity.
Finally, co-segmentation methods assume that the appearance of the background significantly changes from frame to frame. However, as CrowdCam images capture the same event, the background is usually consistent. Moreover, co-segmentation methods do not distinguish between static and dynamic objects.

An alternative to the above mentioned methods is to detect  dynamic regions using  a dense  Structure-from-Motion (SFM) procedure; the static regions will be matched and reconstructed in 3D and the dynamic regions are all the remaining pixels. 

In practice, dense correspondence in CrowdCam images is prone to many errors and holes. For the static regions, the wide baseline causes changes of appearance and occlusions. The moving objects cause additional occlusions (see Fig.~\ref{fig:occlusions}) and  may undergo significant deformations, due to non-rigid motion. This makes it very difficult to reliably match them across images (as demonstrated in Sec.~\ref{sec:results}). A straightforward use of epipolar constraint for distinguishing between dynamic and static regions (e.g., \cite{Luong1996,Yuan2007}), will also suffer from  matching  failures on CrowdCam data.

We propose a novel method for detecting the dynamic regions of a scene from CrowdCam images. Our method avoids 3D reconstruction  and does not rely on establishing dense correspondences between the images. We assume that epipolar geometry can be computed between some pairs of images. 
We treat each image as a reference, and compute a {\em dynamic probability map} that represents the probability of each pixel to be a projection of a dynamic 3D point.
 
\begin {figure*}[t]
	\begin{tabular}{c c c c c}
\includegraphics[width=0.2\linewidth]{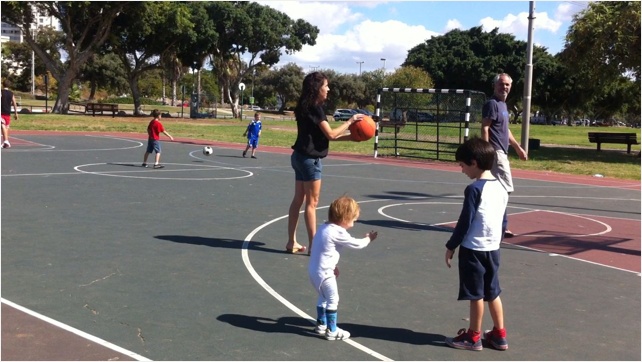} &
	\includegraphics[width=0.2\linewidth]{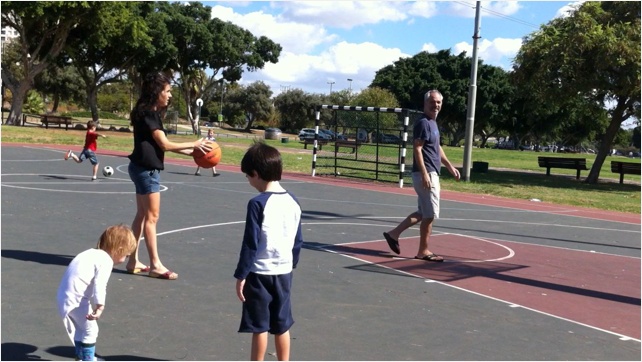} &
	\includegraphics[width=0.2\linewidth]{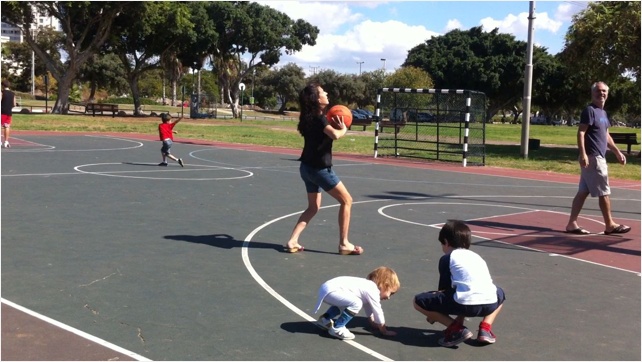} &
	\includegraphics[width=0.2\linewidth]{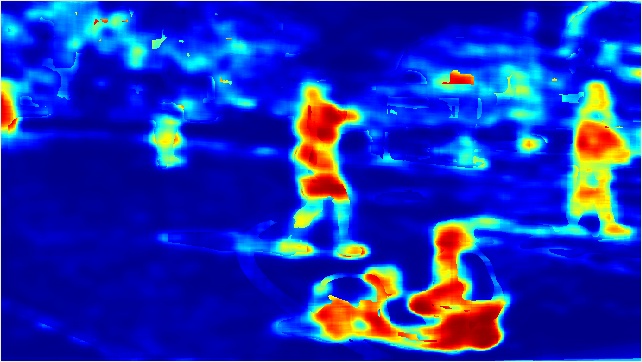} &
	\includegraphics[width=0.2\linewidth]{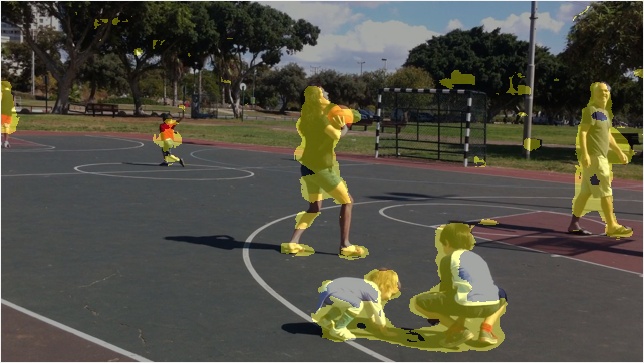} \\
	(a) & (b) & (c) & (d) & (e)\\
	\end{tabular}
	\caption{Three images out of a set of eight: (c) was regarded as the reference image; (a) and (b) are two of the seven support images. The dynamic probability map and a its thresholded map  are presented in (d) and (e), respectively. }
\label{fig:BB_example}
\end {figure*}

An example of a set of three out of eight images, the computed dynamic probability map for one of the images, and a thresholded map are presented in Fig.~\ref{fig:BB_example}. The maps clearly contain information about the dynamic regions.

The method works as follows. First, the dynamic probability map is computed for a reference image and  another (support) image from the set, for which the epipolar geometry can be computed. The probability of a  pixel to be a projection of a general moving 3D point depends on the probability that it has a match in the other image along the corresponding epipolar lines;
projections of  static background must  lie on corresponding epipolar lines.

We do not try to find the correct match, but  instead compute the likelihood that there exists at least one potential match along the epipolar line. In this way we capture the likelihood that a match exists: doing so  decreases the probability that the pixel is {\em dynamic}. This method allows us to deal with matching errors that are due to lack of texture, repeated structure, occlusions,  and so on. To reduce ambiguity, the matching is defined on  epipolar patches, i.e., patches confined by pairs of matching epipolar lines.

Each image may serve as a reference image associated with a subset of support images (for which epioplar geometry can be computed). We then aggregate, for each reference image, the matching probability maps computed using each of its support images, to obtain the final high quality dynamic probability map. This aggregation is necessary because a single map may be unreliable due to accidental matching, resulting in low dynamic probability, or  due to occlusions, resulting in high dynamic probability (see Sec.~\ref{sec:PairOfImages}). However,  these cases are unlikely to consistently repeat w.r.t. all support images. Hence,  the results improve  as the number of support images increases.

The main contributions of this paper are (i) the introduction  of the new problem of detecting dynamic regions  from CrowdCam data; (ii) a voting-based approach that avoids dense correspondence or 3D modeling; (iii) aggregation of information from multiple views for distinguishing between moving objects and occluded regions; (iv) using candidate {\em epipolar patches}  matching while avoiding rectifications between each pair of images (see Sec.~\ref{sec:confined}).

\section {Related Work}
\label{sec:related}

We next review existing methods for detecting moving regions under various setups.

\paragraph{Structure-From-Motion:} SFM methods recover the 3D structure of a static scene \cite{wu2013visualsfm}. As stated earlier, they fail on data such as ours because the images are relatively few and far between. Extensions of SFM methods to handle dynamic scenes, such as non-rigid-SFM \cite{BreglerHB00} are not applicable because they focus on video rate input which is not the case here.

More relevant is the work on temporal SFM from image collections \cite{SchindlerDK07,MatzenS14}. These methods attempt to recover the 3D structure of the scene, as well as organizing the images in their correct temporal order. This is done by analyzing occlusion relationship between points over time and requires sufficient images for proper 3D reconstruction.These methods assume that it is possible to obtain 3D model on which to analyze occlusion relationships which is not always the case, as we show later in the paper. Our approach, on the other hand, sidesteps the need for a 3D reconstruction altogether.

\paragraph{Change Detection:} Change detection algorithms can be done in 2D or 3D.
2D change detection algorithms (also known as background subtraction) are based on comparing a frame to a learned model of the background. This requires an accurate
alignment of several images into the same coordinate frame
(often obtained by using a static camera).  A comprehensive survey
of background subtraction methods is provided in \cite{cristani2010background,radke2005image,lu2004change}.

3D change detection algorithms compare images to a 3D model. For example, Pollard and Mundy \cite{PollardM07} detect changes in a 3D scene observed from an aerial vehicle. They model the scene with a voxel grid model of the scene and determine if the new image was generated by it or not. Similarly, \cite{TanejaBP15} determine city scale change detection by comparing a cadastral 3D model of a city to a 3D model recovered from panoramic images. 

2D Change detection algorithms are not applicable to CrowdCam sets because the scenes are not necessarily flat or distant, and the images cannot be aligned (see supplementary material). 3D Change detection algorithms are not applicable to CrowdCam sets because they rely on a dense 3D reconstruction of the scene and, as we show later in this paper, this is a challenging task because CrowdCam are fairly sparse in space and time.

\paragraph {Motion-based segmentation:}
Motion-based segmentation separates regions in the image that correspond to different motion entities. It usually deals with video sequences.
The classic approach to motion segmentation is based on two-frame optical flow, while recent approaches consider a set of frames and examine the movement characteristics over time \cite{sun2012layered,ochs2014segmentation}.
While early approaches estimate the optical flow and the segmentation independently \cite{wang1994representing,shi1998motion}, optical flow estimation and segmentation were later considered as a joint optimization problem \cite{cremers2005motion,brox2006variational,sun2012layered}. Such methods are not applicable in our case since  no video sequence is available.

Only two papers introduce methods for segmenting motion fields
computed from a  wide-baseline pair of still images, which is  similar to the setup we considered \cite{Wang2015,gullapally2015dynamic}.
The first method is based on  matching feature points, and then  
 minimizing a function that divides the matching  into continuous groups of rigid motions. We, on the other hand, do not assume rigid motion, nor do we assume that correspondence between features of moving objects can be computed.
The second method is based on computing dense correspondence and segment them  into two main motions. The algorithm is limited to regions where dense correspondence can be calculated. We show in Sec.~\ref{sec:results} the limitations of dense correspondence methods on our datasets.

\paragraph {Co-segmentation:}
Co-segmentation is typically defined as the task of jointly segmenting `something similar' in a given set of images (e.g., \cite{rother2006cosegmentation,mukherjee2009half}.  
The problem of co-segmentation is  different from the  one we aim to solve, since it does not distinguish between static and dynamic objects, and it does not take 3D information into account. Hence, it is not suitable to our problem.

\paragraph {Multi-view object segmentation:}
Algorithms of this family address the task of unsupervised multiple image segmentation of a single physical object, possibly moving, as seen from two or more calibrated cameras. The input may either be still images or video sequences. 
Zeng {\em et al.} \cite{gang2004silhouette} coined the problem, and proposed an initial rudimentary silhouette-based algorithm for building segmentations consistent with a single 3D object. Many methods follow this initial trend by building explicit 3D object reconstructions and alternating with image segmentations of the views based on foreground/background appearance models~\cite{campbell2010automatic,guillemaut2011joint}.

Our method avoids the 3D reconstruction. Recovering the 3D structure of a dynamic scene often requires prior knowledge about the 3D structure or the motion of objects, and a very large number of images, which we do not assume to have.
The limitations of dense 3D reconstruction on our data are presented in detail in Sec.~\ref{sec:results}.

Another line of related work  is that of objectness proposal, where the goal is to suggest image windows that are likely to contain an object of interest (e.g., \cite{AlexeDF12}). 
Since the input to these methods is  a single image, they cannot reason about motion information, which often indicates the interesting regions. Our method may  be integrated into objectness proposal algorithms, in addition to other single image cues such as saliency, color contrast and edge density.

\section {Method}
\label {sec:method}

We are given a set of $n$ images, taken by various uncalibrated cameras. We assume that for each image we can compute its epipolar geometry w.r.t. a subset of  images, termed \emph{support set}. This assumption holds when there are sufficient static features in the set of images and the dynamic features are treated as outliers by a RANSAC algorithm  which is used to compute the epipolar geometry (e.g., \cite{goshen2008balanced}). For each image, we compute a matching probability map based on its epipolar geometry with each of its support images and then merge all those maps into a dynamic probability map for that image. We next describe a method to compute a matching probability map from a pair of images, and then discuss the aggregation of these maps to compute a dynamic probability map.

\subsection{Pair of images}
\label{sec:PairOfImages}
\begin{figure}[t]
\begin{tabular}{@{}cccc@{}}
\includegraphics[width=0.33\linewidth]{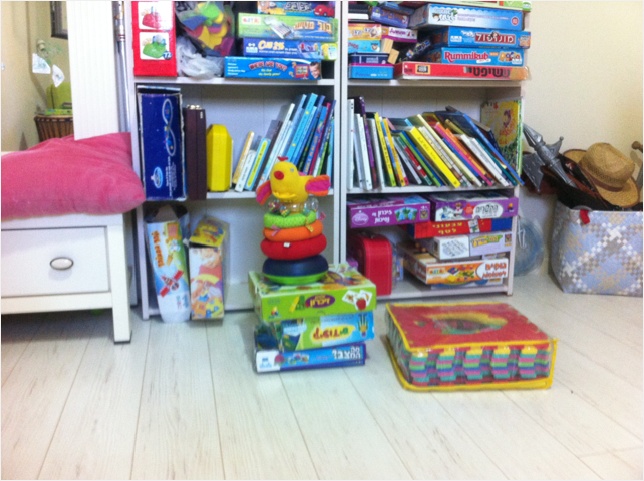}&
\includegraphics[width=0.33\linewidth]{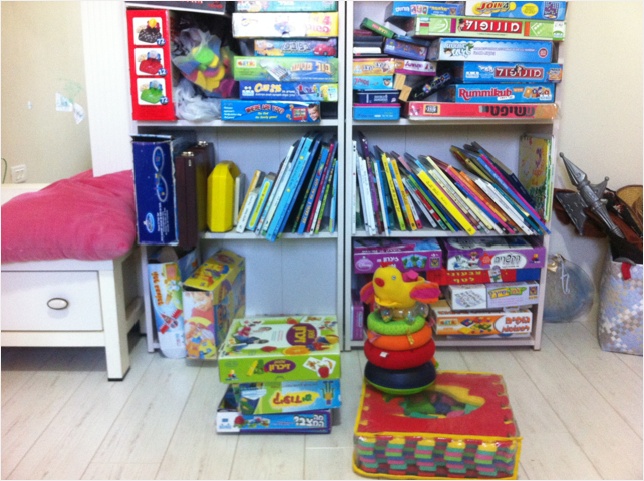}&
\includegraphics[width=0.33\linewidth]{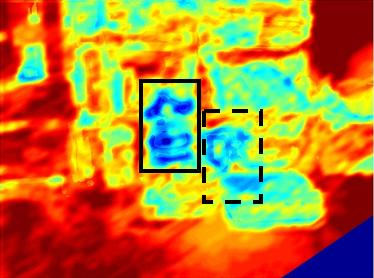}\\
(a) & (b) & (c) \\
\end{tabular}
\caption{
Dynamic object shadow: (a) a reference image;  (b) a support image; (c) the computed static probability map. Two low probability regions are depicted: the true location of the moving object (a continuous rectangle), and the moving object's shadow (dashed rectangle).
}
\label{fig:shadow}
\end{figure}

Given a reference image $I$ and a single support image,~$I_s$, we compute ${P(\textbf{x}|I_s)}$, the probability that a pixel, $\textbf{x}\in I$, is static and non-occluded. 
Observe that ${P(\textbf{x}|I_s)}$ is low not only for pixels in dynamic regions, but also for pixels in the following regions: 
(i) Out of field of view; (ii) Occluded due to different viewpoints; (iii) Occluded by the moving object in $I_s$, e.g., Fig.~\ref{fig:shadow}(c) (we refer to these regions as {\em dynamic object shadows});
(iv) Regions for which the descriptor fails to detect the similarity due to  variations in appearance. Such variations exist due to the change of viewpoint and/or illumination, and the difference between the cameras' inner parameters.

It is evident from this list that failure to find a match does not necessarily mean that the pixel belongs to a dynamic region. As we show later, when using  many support images, the probability of finding matches for static pixels will be significantly higher than for dynamic pixels (e.g.,  Fig.~\ref{fig:set_power}).

\subsubsection{The set of  epipolar patches}
\label{sec:confined}
Matching a single pixel is very noisy and we work with patches instead. The probability that a pixel $\textbf{x}$ has a match is derived from the probability that each of the patches covering $\textbf{x}$ has a match.

When building a set of candidate pairs of patches for correspondence, the first step in the calculation is to define the patch's shape and size in $I$ and in~$I_s$. 

\begin {figure}[t]
	\begin{center}
	\begin{tabular}{c c c}
			\includegraphics[height=0.29\linewidth]{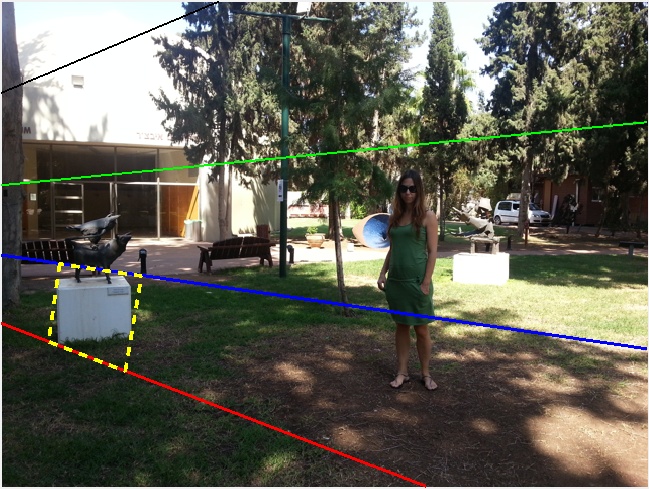}&
		\includegraphics[height=0.29\linewidth]{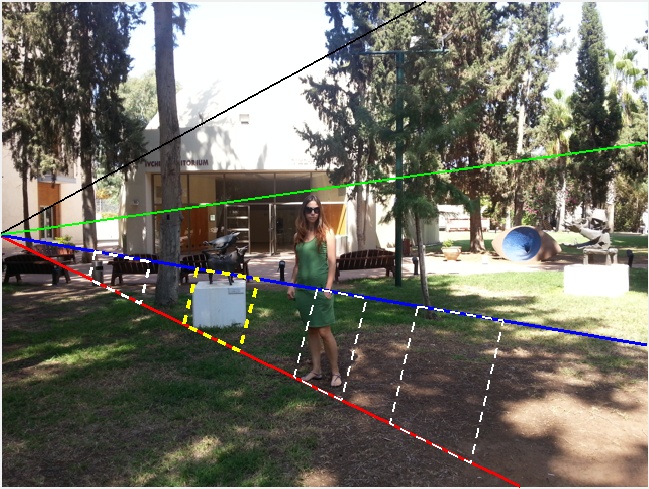} &
		\includegraphics[height=0.29\linewidth]{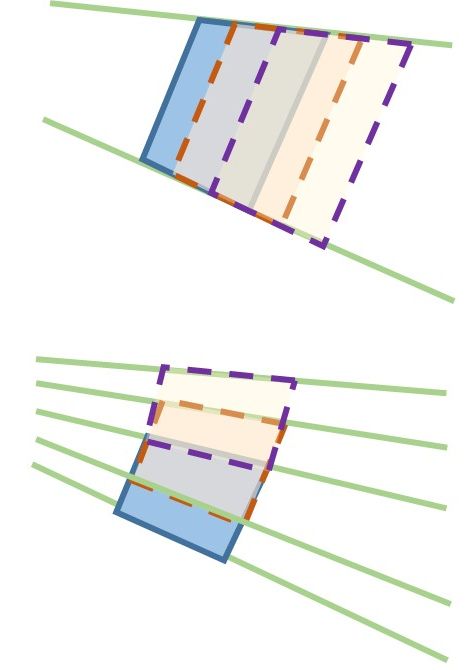}\\
		(a) & (b) & (c)
		\end{tabular}
			\end{center}
	\caption{ 
	(a)-(b) The corresponding region of the statue's base (the white object to the left of scene) is located between the corresponding pair of epipolar lines. The width of the corresponding patches differs between the two images. Some additional possible matching patches of varied sizes are depicted for illustration purposes. (c) Overlapped patches defined between a pair of epipolar lines (green lines) and  across epipolar lines. }
\label{fig:el}
\end {figure}

\begin {figure*}[t]
	\begin{tabular}{c c c c c c c}
		\includegraphics[width=0.0109\linewidth]{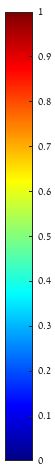}&
		\includegraphics[width=0.16\linewidth]{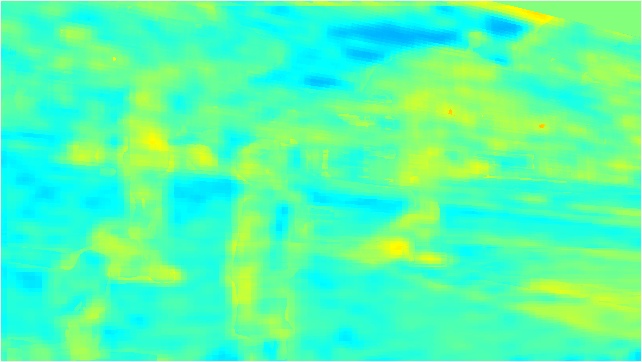}&
		\includegraphics[width=0.16\linewidth]{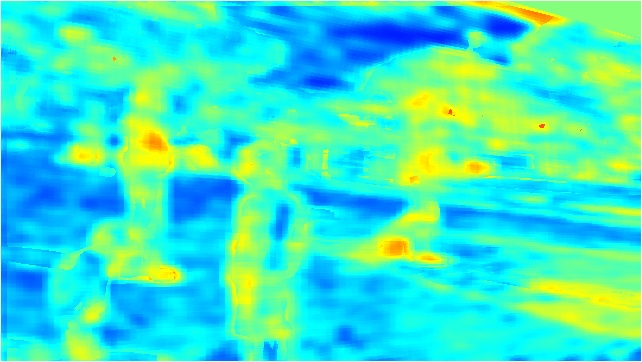}&
		\includegraphics[width=0.16\linewidth]{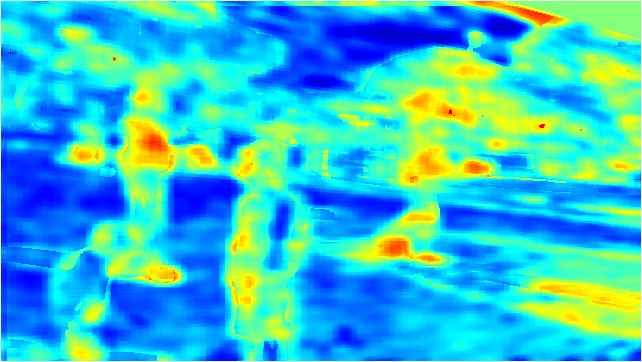}&
		\includegraphics[width=0.16\linewidth]{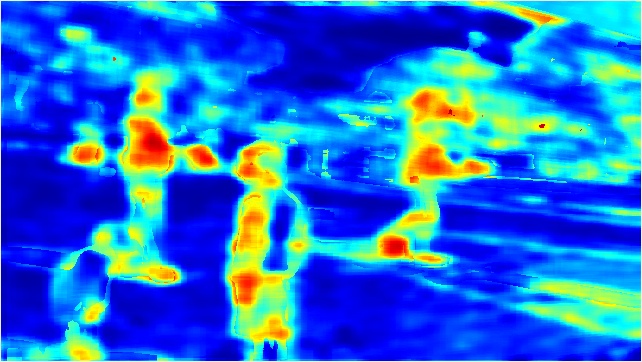}&
		\includegraphics[width=0.16\linewidth]{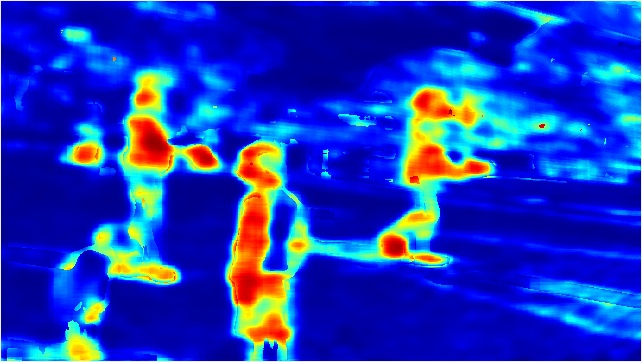}&
		\includegraphics[width=0.16\linewidth]{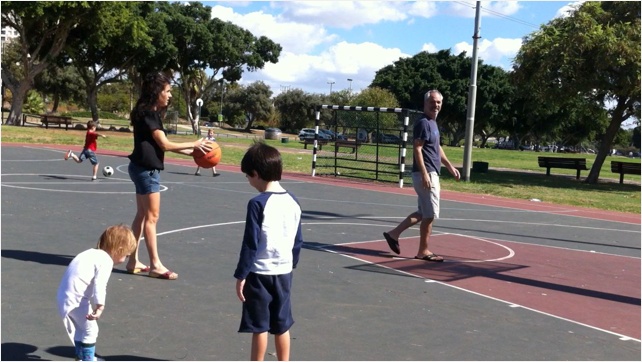}\\	
			 & (1) & (2) & (3) & (5) & (7) \\
	\end{tabular}
	\caption{The improvement of the dynamic probability map as a function of the number of support images. The number of support images is depicted beneath the probability maps; the color bar, which is common to all the probability maps, is depicted to the left.}
\label{fig:set_power}
\end {figure*}

Rectangular patches are commonly used, but they are more appropriate for rectified pairs. 
Since in the general case the epipolar lines are not parallel, each of the possible matches may be of different height. 
We consider  patches that are confined between  pairs of epipolar lines -- {\em epipolar  patches}.
The correspondence of a static epipolar patch in $I$ is an epipolar patch in~$I_s$, confined by the pair of corresponding epipolar lines. This follows directly from epipolar geometry of static regions. 
The use of epipolar patches determines the height of the candidate patches in $I_s$, for matching.  The ambiguity regarding the candidate patch width remains, since the scale of the object in $I_s$ is unknown (see Fig.~\ref{fig:el}).

In practice, we compute a set of epipolar lines in the reference image,  and a set of patches between each pair of adjacent lines is defined, with up to $2/3$ overlap between them (see Fig.~\ref{fig:el}(c)).  
In a similar manner, the candidate  set of patches is computed in the support image between the corresponding epipolar lines but with 3 different widths.  The epipolar lines are parametrized by the angle of the line where the epipole is taken to be the origin.  For obtaining overlap across epipolar lines, additional epipolar lines are considered
with  $1/3$ and $2/3$ shift of the angle (see Fig.~\ref{fig:el}(c)).
Thanks to the use of overlapping patches, the confidence that a pixel is static is measured a few times, and the final probability map is smoother and more robust.

\subsubsection {Patch confidence measure}
Let ${C(\textbf{r}|I_s,\theta)}$ be the confidence that the patch $\textbf{r}\in I$ is a projection of a static scene region not occluded in $I_s$.
This confidence is based on the similarity between~$\textbf{r}$ and its nearest neighbor among~${\cal R'}$, the set of candidate matching patches in $I_s$. Formally, 
\begin{equation}
C(\textbf{r}|I_s, \theta) = max_{\textbf{r'}\in {\cal R'}}\{sim_\theta(\textbf{r},\textbf{r'})\},
\label{eq:conf}
\end{equation}
where $sim_\theta(\textbf{r},\textbf{r'})$ is the similarity between the two patches, using the descriptor~$\theta$ (e.g., HOG or color histogram). 
The confidence is normalized to the range $(0,1)$ for each descriptor  by mapping the range of  $C(\textbf{r}|I_s, \theta)$ of  all pairs of reference and support images. We denote the normalized value by $\hat C(\textbf{r}|I_s, \theta)$.

Albeit simple, this measure turns out to have important and nontrivial values.
Ambiguity often makes it difficult to choose
 the best candidate. For example, when the background is periodic or uniform, there may be more than a single patch with high correspondence confidence along the epipolar line. 
As we do not aim to recover the 3D structure, 
locating the correct match is not important for the success of the algorithm. We merely focus on the question of whether or not a good correspondence exists.
Clearly, if the best match has low confidence, the pixel is unlikely to be a projection of a non-occluded static 3D point.

Extensive research exists regarding the difficulties of choosing the best descriptor of a patch and the best method of computing similarity between descriptors of two patches. As expected, we  found that the optimal descriptor depends on the image set -- the extent to which the image colors change, and the various textures of the captured objects. The algorithm proposed in this paper may be used with any set of descriptors and similarity measures.

\subsubsection{Matching probability map}
We treat the confidence as a probability and use it to build a probability map that holds,  for each pixel $\textbf{x}$ in $I$, the  probability that~$\textbf{x}$ is static and not occluded, ${P(\textbf{x}|I_s)}$. This probability measure is based on the confidence that a good match of the pixel's region exists along the epipolar line, as described in Eq.~\ref{eq:conf}.  

Let ${\cal R} _x$ be the set of patches that contain a pixel $\textbf{x}$ and let~$\Theta = \{\theta_1 ... \theta_m\}$ be the set of descriptors.
The matching probability  of a pixel, ${P(\textbf{x}|I_s)}$, is calculated as the weighted expectation estimation of the set of confidences computed for each of the patches in  ${\cal R} _x$ with each  descriptor in~$\Theta$. 
It is given by: 
\begin{equation}
P(\textbf{x}|I_s) = \sum_{\theta\in {\Theta},\textbf{r}\in {\cal R} _x} w_r w_\theta \hat C(\textbf{r}|I_s, \theta).
\label{eq:pair}
\end{equation}
Here $w_\theta$ and $w_r$ are the  weights of the confidence of a descriptor~$\theta$ and the location of $\textbf{x}$ within the patch, respectively. The value  $w_\theta$ is  predefined by the user for each descriptor. We set  $w_r$ to be inverse proportional to  the distance, $d(\textbf{x},\textbf{r}_c)$,  of the pixel  from  the patch center, $\textbf{r}_c$. In our implementation,  $w_r=e^{-{d(\textbf{x},\textbf{r}_c)^2} / {2 \sigma^2}}$  and  
 $\sigma =  {max_{x,r_c}\{d(\textbf{x},\textbf{r}_c)\}} /{3}$, where the max is taken over all patches in all of the images.
Weights are normalized to sum to one for each pixel; therefore $P(\textbf{x}|I_s)$  is guaranteed to be in the range $[0,1]$, and we can regard it as probability. 

\subsection{A set of images}

Combining the results obtained from multiple support images is analogous to considering the testimonies of a few witnesses who viewed the same scene from different locations. Regions that are occluded or out of view in one image are expected to be visible in other images  (see Fig.~\ref{fig:occlusions}). Similarly, if the motion coincides with the epipolar lines in a pair of images, it is unlikely to coincide with the epipolar lines with respect to the other images. 
Fig.~\ref{fig:set_power} illustrates the effect of using an increasing number of support images.

Our goal is to compute the dynamic probability, $P(\textbf{x})$, given a set of support images  $\{I_{\cal S}\}$. We compute the matching probability, $P(\textbf{x}|I_s)$, for each $I_s\in I_{\cal S}$, and combine the probabilities as follows:

\begin{equation}
P_{static}(\textbf{x}|I_{\cal S}) = \cfrac{ \underset{s\in {\cal S}} {\Pi} {P(\textbf{x}|I_s)}}{\underset{s\in {\cal S}} {\Pi}{P(\textbf{x}|I_s)}+ {\underset{s\in {\cal S}} {\Pi} {(1-P(\textbf{x}|I_s))}}}.
\label{eq:set}
\end{equation}
$P_{dynamic}(\textbf{x})$ is the complementary probability.
Note that the above aggregation of probabilities has the following characteristics: the aggregated probability of a few probabilities that are higher than 0.5 is higher than each of the input probabilities. Similarly, when all of the probability values are lower than 0.5, the combined probability measure is lower than each of the inputs. An input probability of  0.5 does not influence the combined probability -- in this case the combined probability is determined by the rest of the input probabilities. Moreover, high and low probabilities balance each other out and result in  a probability that lies in between them.
Before combining the probabilities we add a preliminary step of remapping the probability values to the range (0.3,0.7), to avoid the overinfluence of extreme values of $P(\textbf{x}|I_s)$ (e.g., 0 or 1). 
The use of multiple images,  multiple overlapping patches, and descriptors per pixel, allows our method to handle false correspondences, 
as we demonstrate in the next section.

\begin {figure}[t!]
	\begin{tabular}{c c c }
			\includegraphics[width=0.33\linewidth]{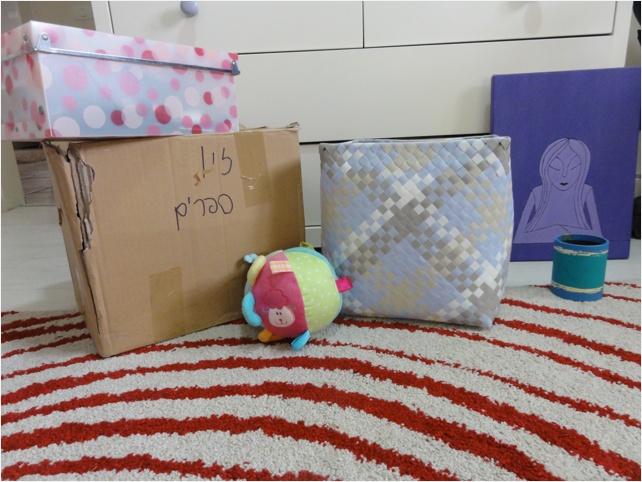}&
		\includegraphics[width=0.33\linewidth]{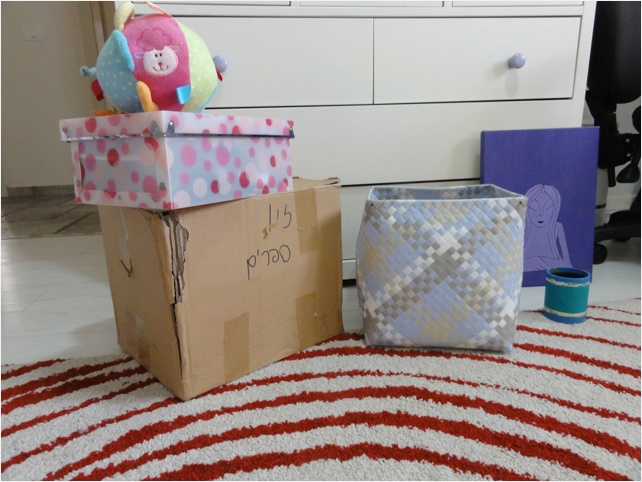}&
		\includegraphics[width=0.33\linewidth]{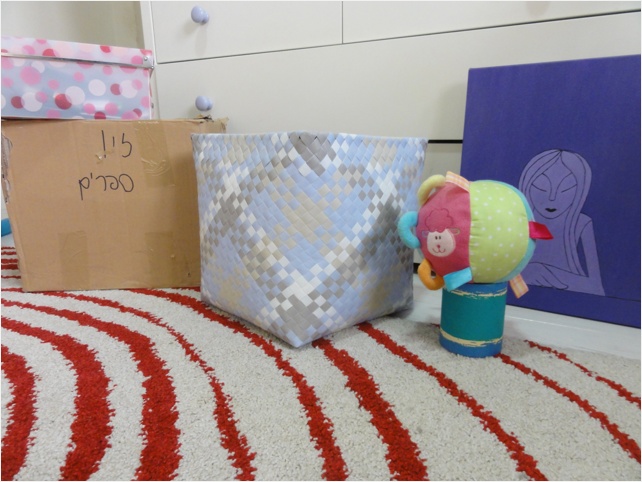}\\		
		\includegraphics[width=0.33\linewidth]{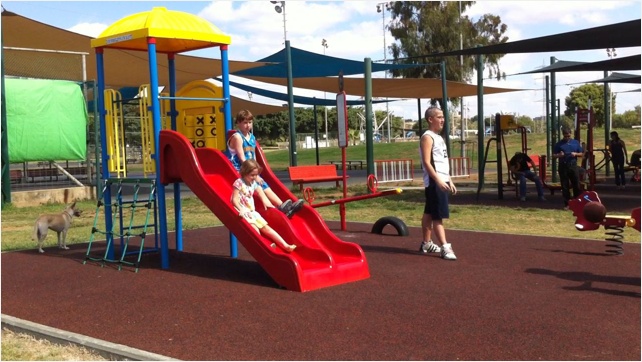}&
		\includegraphics[width=0.33\linewidth]{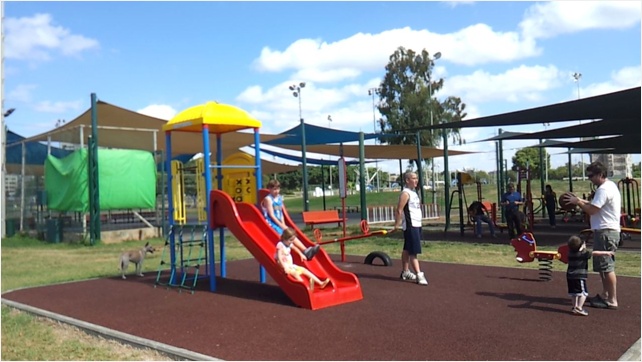}&
		\includegraphics[width=0.33\linewidth]{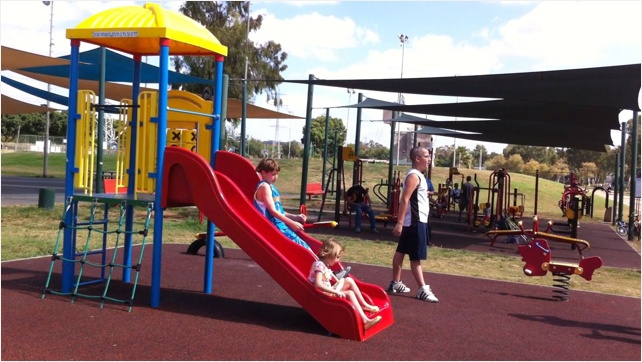}\\		
		\includegraphics[width=0.33\linewidth]{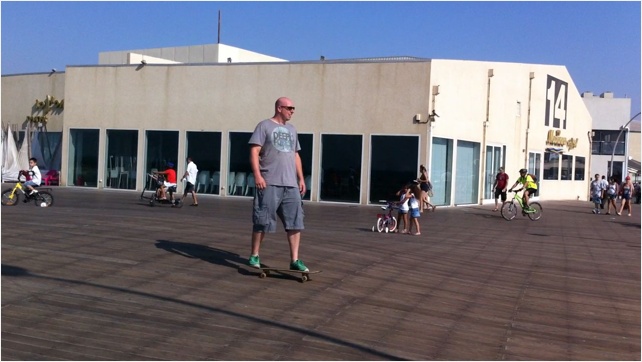}&
		\includegraphics[width=0.33\linewidth]{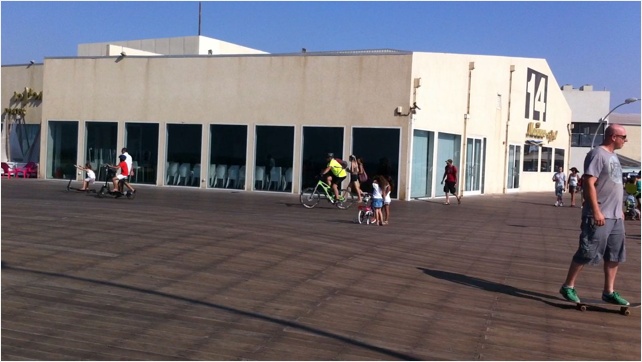}&
		\includegraphics[width=0.33\linewidth]{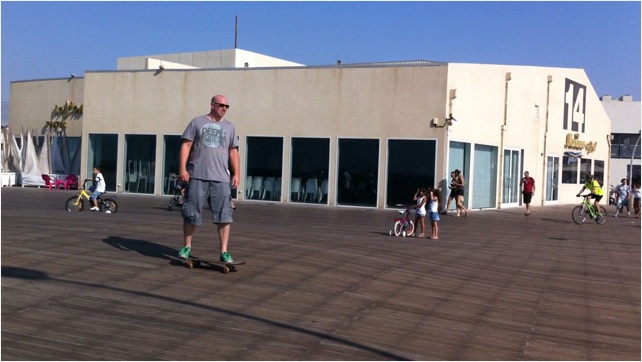}\\	
					\includegraphics[width=0.33\linewidth]{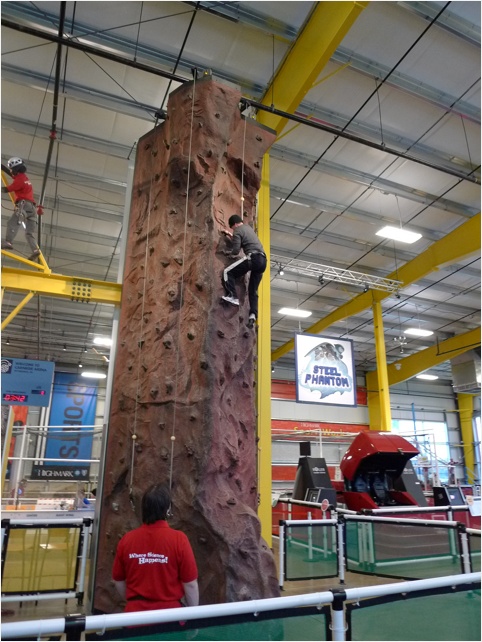}&
		\includegraphics[width=0.33\linewidth]{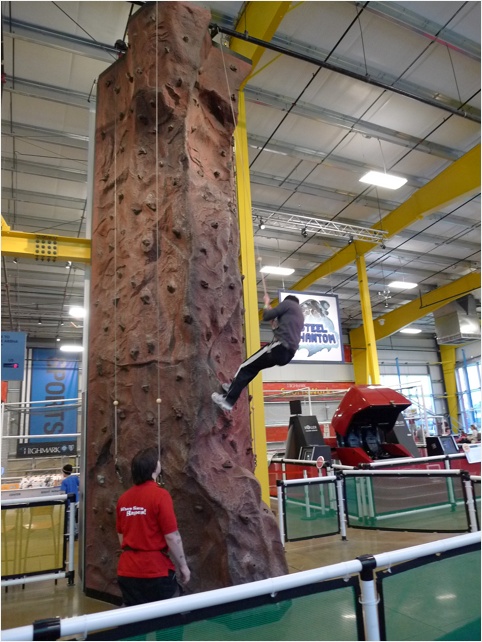}&
		\includegraphics[width=0.33\linewidth]{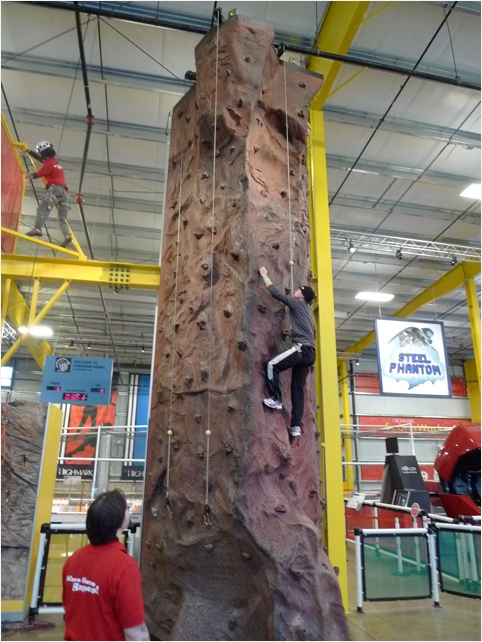}\\	
	 \end{tabular}
	\caption{Three images of each dataset. From top to bottom:  Toy Ball, Playground, Skateboard, Climbing. (The helmet and basketball sets can be viewed in Fig.~1 and Fig.~\ref{fig:BB_example}, respectively).}
\label{fig:data_sets}
\end {figure}

\begin {figure*}[!t]
\begin{center}
	\begin{tabular}{c c c }
		\includegraphics[width=0.25\linewidth]{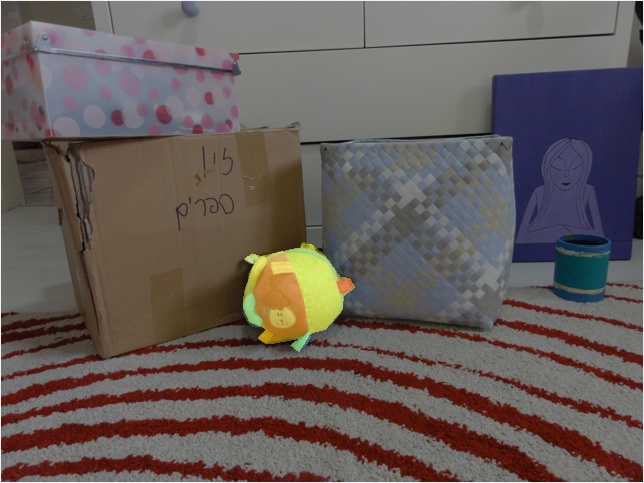}&
		\includegraphics[width=0.25\linewidth]{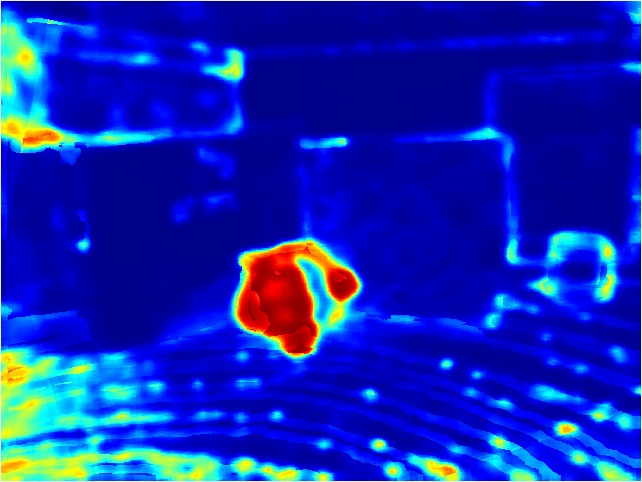} &
		\includegraphics[width=0.25\linewidth]{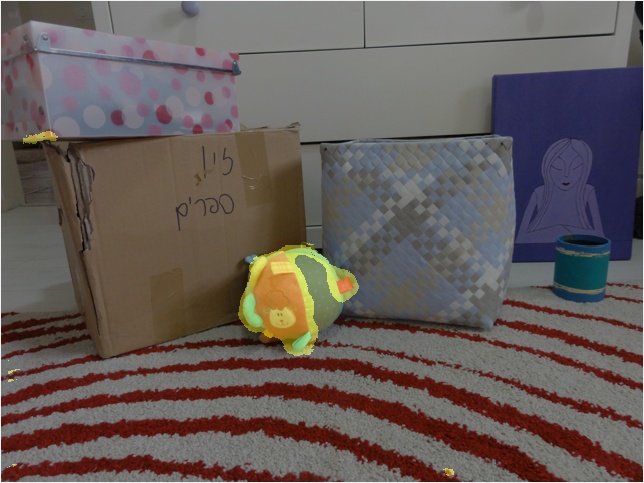} \\ 
		\includegraphics[width=0.25\linewidth]{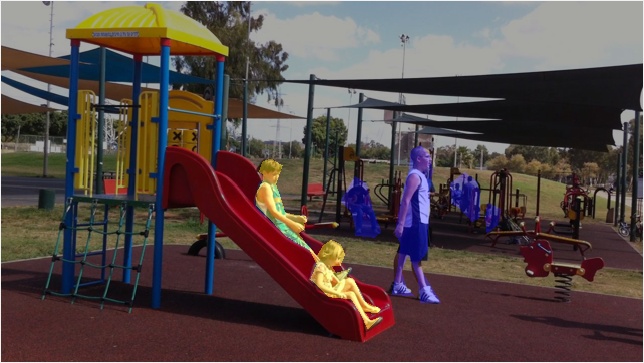}&
		\includegraphics[width=0.25\linewidth]{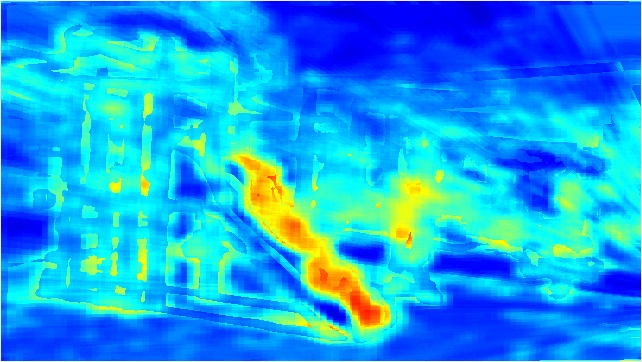}&
		\includegraphics[width=0.25\linewidth]{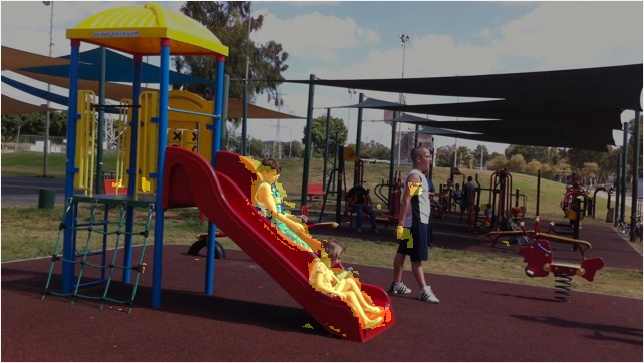}\\	
		\includegraphics[width=0.25\linewidth]{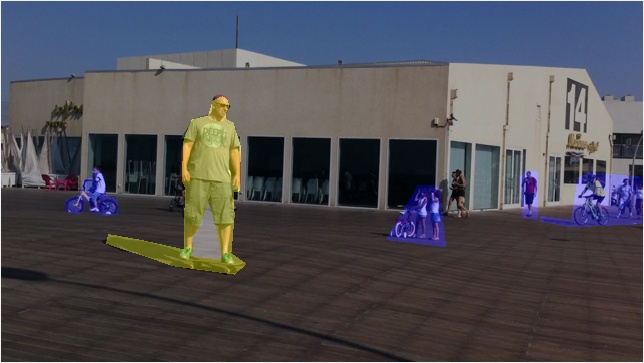}&
		\includegraphics[width=0.25\linewidth]{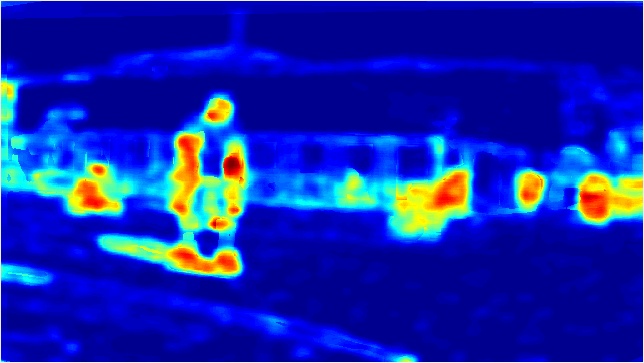}& 
		\includegraphics[width=0.25\linewidth]{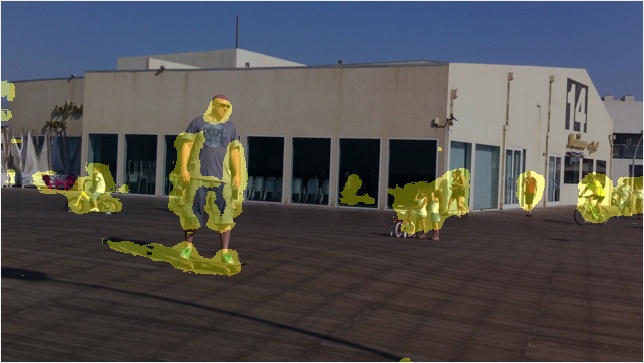}\\
		\includegraphics[width=0.25\linewidth]{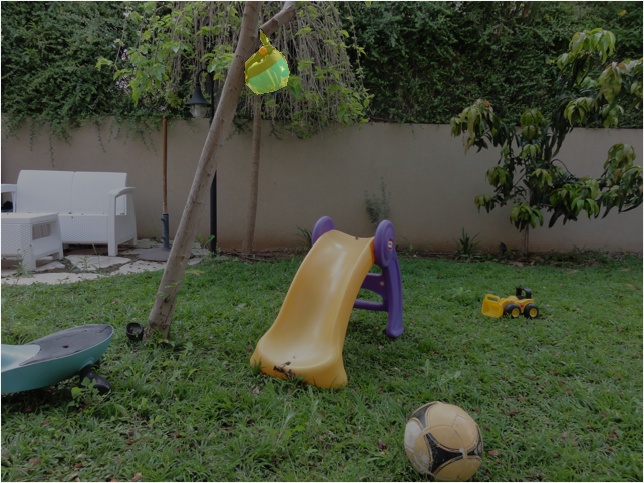}&
		\includegraphics[width=0.25\linewidth]{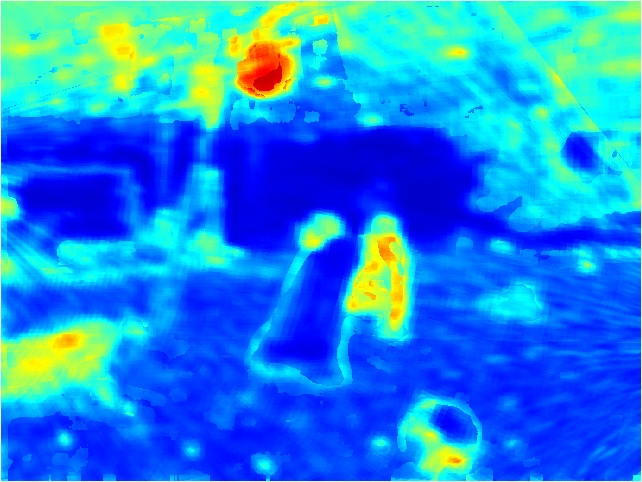}&
		\includegraphics[width=0.25\linewidth]{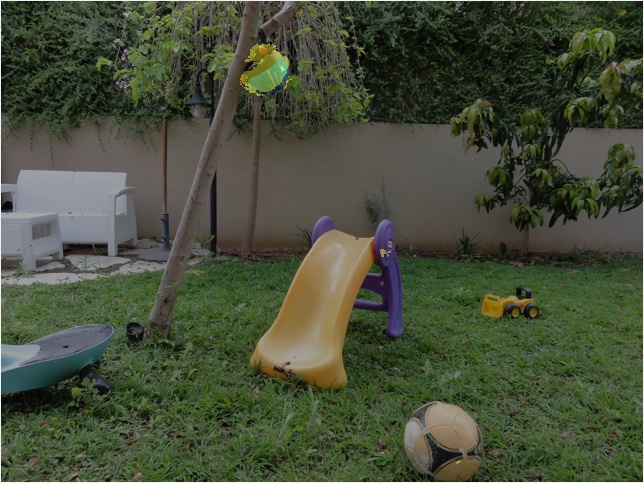}\\  	
		\includegraphics[width=0.25\linewidth]{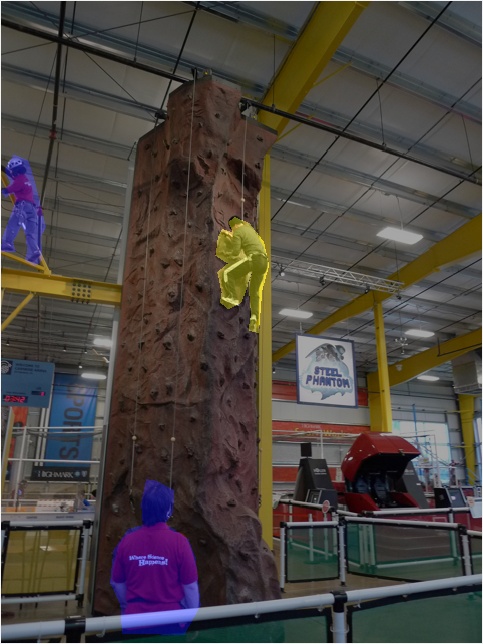}&
		\includegraphics[width=0.25\linewidth]{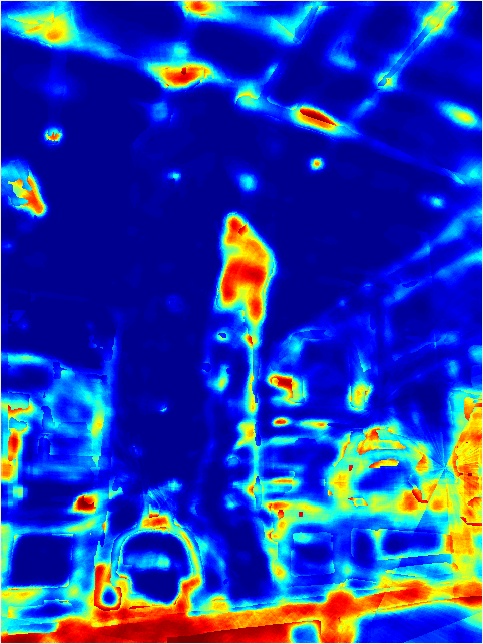}&
		\includegraphics[width=0.25\linewidth]{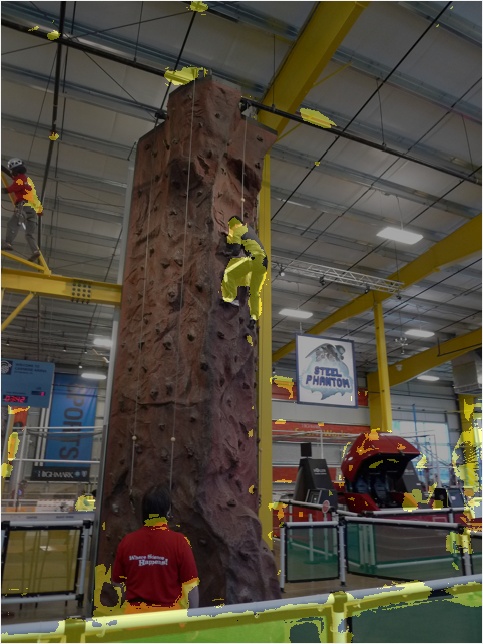} \\		
	 \end{tabular}
	\end{center}
	\caption{From left to right: ground truth mask, dynamic probability map, and the thresholded map, for each of the datasets. The `don't care' areas in the ground truth masks are marked in blue. Note that from the thresholded map, all dynamic regions are detected, only the last row contains many false positive detections.}	
\label{fig:results}
\end {figure*}

\begin {figure*}[t]
\centerline{
	\begin{tabular}{c c c c}
		\includegraphics[height=0.18\linewidth]{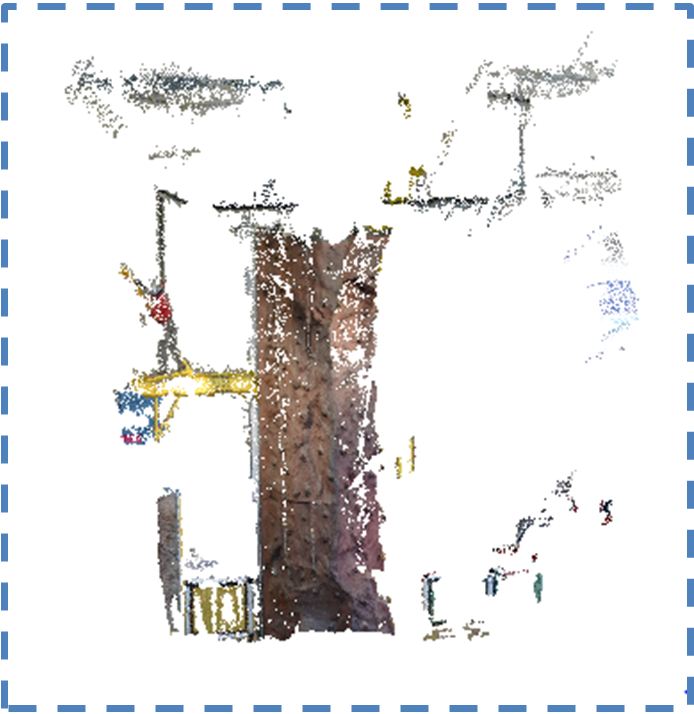}&
		\includegraphics[height=0.18\linewidth]{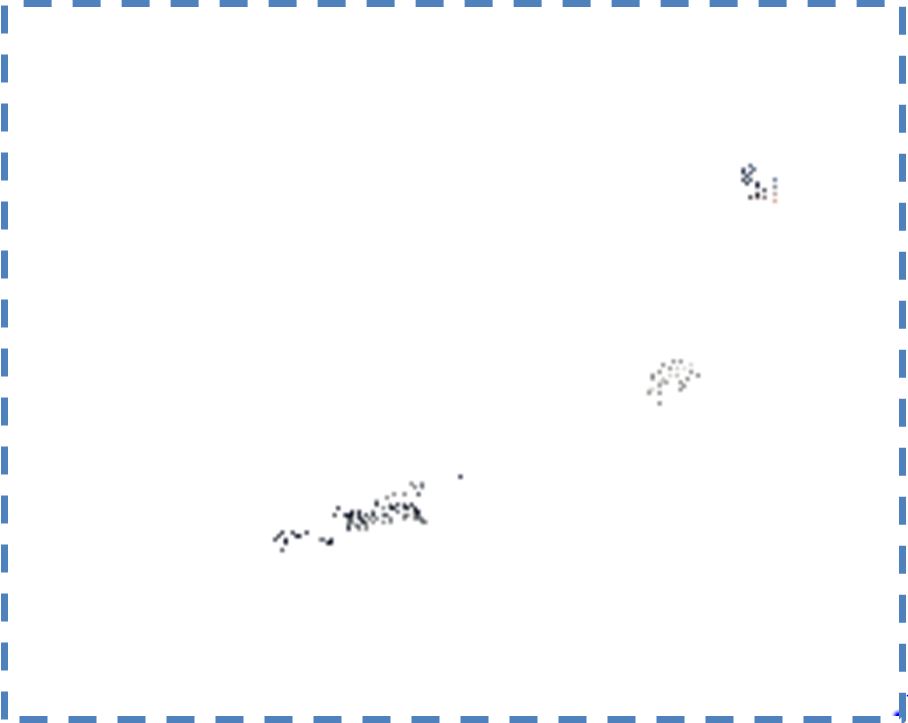}&
		\includegraphics[height=0.18\linewidth]{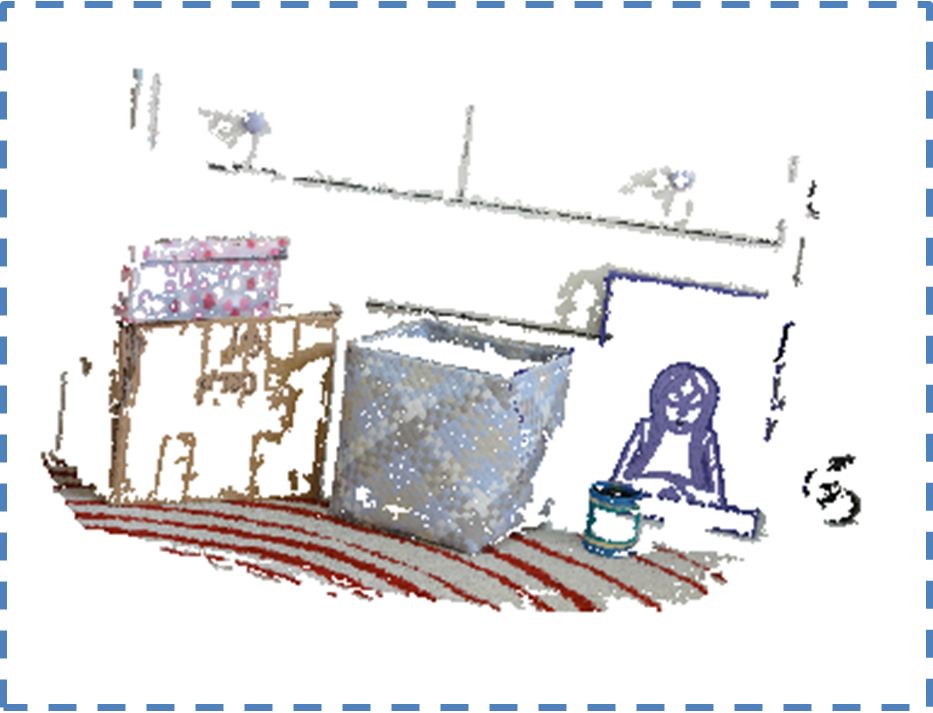}&
		\includegraphics[height=0.18\linewidth]{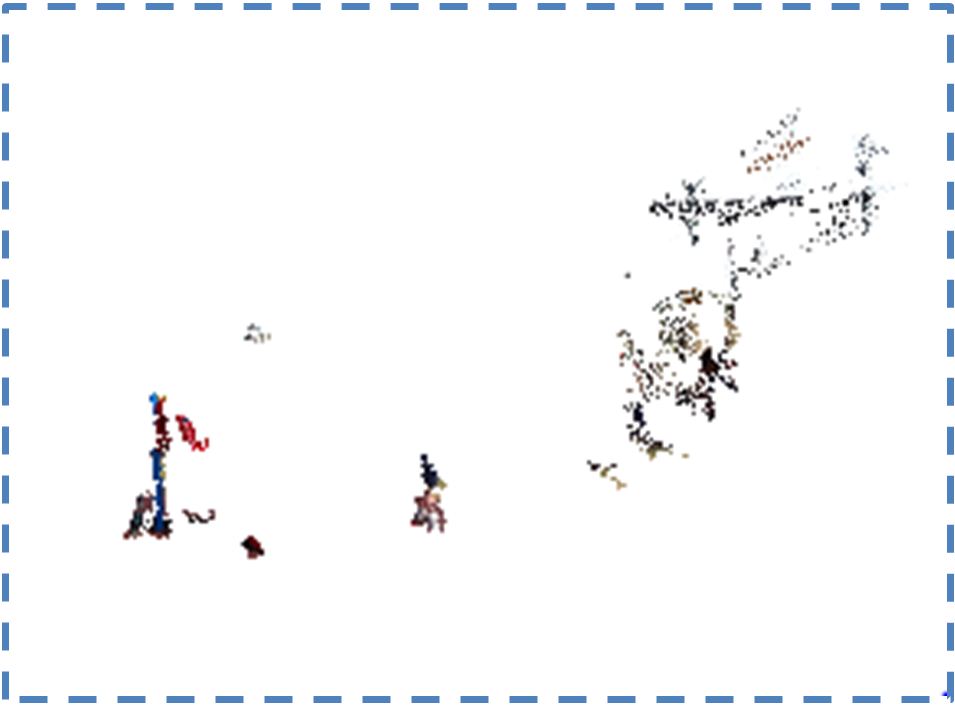}\\		
	\end{tabular}}
	\caption{ Visual SFM (\cite{wu2013visualsfm}) results on the (left to right) Climbing, basketball, toy-ball and playround sets; the algorithm failed on the skateboard and helmet sets }
\label{fig:sfm}
\end {figure*}

\section{Results}
\label {sec:results}

We implemented the proposed algorithm in MATLAB and tested it on  challenging real-world data sets. (Standard datasets for this task are not available.)

\paragraph{Datasets:}
Three images of each set are depicted in Fig.~1, Fig.~\ref{fig:BB_example}, and Fig.~\ref{fig:data_sets} (the full sets can be found in the supplementary material). The sets capture both indoor and outdoor scenes, single as well as multiple moving objects, and rigid as well as non-rigid (person) objects.
The rock-climbing set was captured by Park {\em et al.} \cite{park20103d}, the playground, basketball and skateboard sets were captured by Basha {\em et al}. \cite{basha2012photo}, and the other two were captured by us.
All images were captured from different viewpoints, without calibration or a controlled setup.
We used the same camera in four of the six image sets to focus on the behavior of the algorithm and not the sensitivity of the descriptors to camera change.

\begin {figure}[b!]
	\begin{tabular}{c c }
		\includegraphics[width=0.5\linewidth]{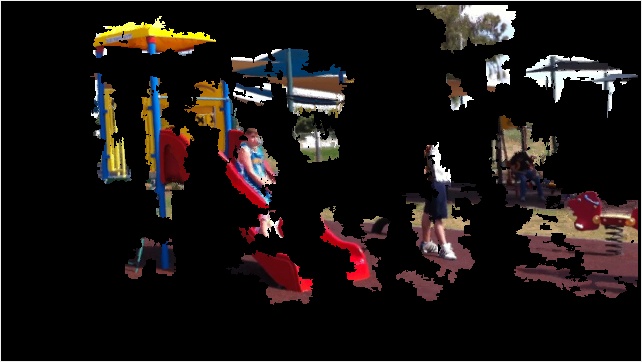}&
		\includegraphics[width=0.5\linewidth]{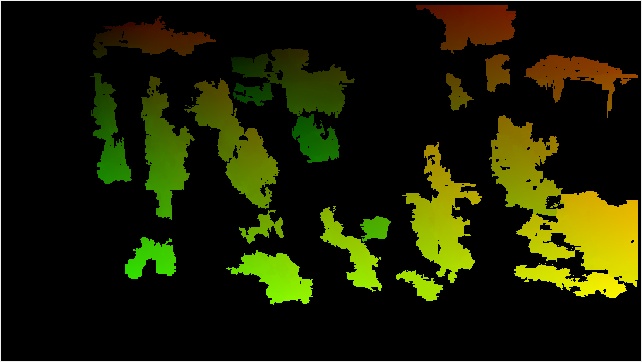} \\
			
		(a) & (b) 	\end{tabular}
	\caption
	[NRDC performance on the playground dataset] 
	{Failure on NRDC: (a) The region for which the NRDC algorithm found matching pixels on a pair of images from the playground dataset,  (b) the  computed confidence map of matching  (Black regions: no matching pixels were found).}

\label{fig:NRDC_res}
\end {figure}

\paragraph{Implementation details:}

 We computed the fundamental matrices of the image pairs using the BEEM algorithm 
\cite{goshen2008balanced},   
and  used only pairs of images where BEEM succeeded.  
The sets of patches in the reference image were chosen such that each pixel was covered by nine patches -- three overlapping patches along the epipolar line, and three across epipolar lines.
We used the same combination of two descriptors for all experiments: a histogram of oriented gradients (HOG) descriptor, and a~2D histogram of the H and S channels of the HSV color representation. The weights of the descriptors were set to~2 and~1, respectively. The similarity of the HOG descriptors was computed using the cosine distance. The similarity of two 2D histograms, $B_1$ and $B_2$, was computed using their intersection over union measure (in our implementation we used $10$ bins per channel).

\subsection{Qualitative Results}
The dynamic probability maps are presented for each  of the datasets as a heat map (blue for static and red for dynamic). We consider independently each image in each dataset as a reference image. Fig.~\ref{fig:results} shows an example of a dynamic probability map for one reference image per set, and its thresholded  map overplayed on the image (more example are presented in the supplementary material).

Overall, we observe that our algorithm successfully assigns high probabilities to the moving regions, in most cases.  Hence, it can be used to detect the dynamic regions.
Observe that these regions are indeed the interesting parts of the scene and hence our method can be used to direct the attention of higher level algorithms to these regions.

\begin {table}[t!]
\begin{center}
	\begin{tabular}{ | l | c | c | l | l | }
	\hline
		\textbf{Image set} & \textbf{Set } &  \textbf{Ave. size of } & \textbf{~~~~Jaccard} &  \textbf{~~~~Jaccard} \\ 
		          &  \textbf{size}  & \textbf{support sets} & opt.~per image & ~~opt.~per set   \\ \hline
		Helmet & 4 & 2 & ~~0.53 $\pm 0.18$~~ & ~~0.36 $\pm 0.28$~~ \\ \hline 	
		Skateboard & 5 & 4 &~~ 0.44 $\pm 0.1$~~ & ~~0.42 $\pm 0.1$~~  \\ \hline
		Playground & 7 &~~2.6~~ & ~~0.37 $\pm 0.11$~~ & ~~0.32 $\pm 0.01$~~  \\ \hline	
		Toy Ball & 7 & 4.5 & ~~0.63 $\pm 0.03$~~ & ~~0.6 $\pm 0.05$~~  \\ \hline
		Basketball & 8 & 7 & ~~0.48 $\pm 0.04$~~ & ~~0.47 $\pm 0.04$~~  \\ \hline
		Climbing  & 10 & 9 & ~~0.15 $\pm 0.05$~~ &~~ 0.13 $\pm 0.04$~~ \\ \hline
				\end{tabular}
				\end{center}
		\caption
		[results - Jaccard]
		{Quantitative Results: for each data set we show the number of images in the set, the average size of the support set for a reference image (images for which computing the fundamental matrices was successful), and the Jaccard measure (higher is better) with a threshold optimized per image and per set.}
\label{tab:results}
\end{table}

In some places the algorithm struggles. This is usually because some of our underlying assumptions are not met in practice. In the skateboard set, the rider's shirt resembles the color of the right windows, and in the  toy-ball set part of the ball is not detected since it resembles parts of the background.
In the challenging climbing set, the man wearing the red shirt at the bottom of the images hardly moves;  therefore only the edges of his silhouette are detected (Fig.~\ref{fig:results}, last row).
The colors of the climber's shirt resemble the colors of some areas of the climbing wall,  and  the shirt detection is weak as a result. 
False positives occur when the descriptor fails to detect similarities. For example,
the matching fails on reflective, transparent and narrow objects (less than patch's width) in the climbing set.

\subsection{Quantitative Results}
We evaluate our method using the Jaccard measure (intersection over union) on manually labeled moving regions. The measure requires a binary map so we threshold the probability map to obtain one. Inspired by the evaluation methodology of the Berkeley Segmentation Data Set \cite{MartinFTM01}, we use two thresholds -- a threshold that optimizes the Jaccard measure of each image, and one that optimizes the mean Jaccard measure of all of the images in a given set.
Examples of manual ground truth masks are shown on the left column of Fig.~\ref{fig:results}, and examples of masks that resulted from thresholding the dynamic probability map are shown on its right column. 

The algorithm was applied to each of the images in the sets and the mean Jaccard measures per dataset are presented in Table~\ref{tab:results}. 
The measure is high for the toy ball, basketball and helmet sets. It is low for the challenging climbing set, as discussed earlier.

\subsection{Comparison to other methods}  
We compared our results to  \cite{gullapally2015dynamic} on the skateboard dataset (see supplementary material), and found that we
 outperforms it by a large margin. 
Our algorithm not only detects all moving regions in the image (rather than only one), but also creates a smoother and more complete region of the skateboard rider.

The reported Jaccard measures by~\cite{gullapally2015dynamic} on a pair of images from the set were 0.37 and 0.28.  
Our results were 0.42 and 0.59, respectively, when only a single support image was used, and 0.45 and 0.6 when 4 support images were used. (For a fair comparison,  we generated a ground truth mask of only the main moving object since  \cite{gullapally2015dynamic} assumes only a single moving object.)
 
In addition, we tested the performance of NRDC 
 \cite{hacohen2011non} on our data sets as a way to evaluate the difficulty of establishing correspondences that
are a necessary first step in the method of~\cite{gullapally2015dynamic}   in particular and SFM algorithms in general.
An example of the performance of the NRDC algorithm on an image of our CrowdCam images is demonstrated in  Fig.~\ref{fig:NRDC_res}, with additional  examples  in the  supplementary material.
Quantitative results show that for the skateboard dataset, NRDC found correspondences for 75\% of the image, but for the rest of the datasets as few as  33\% of the correspondences were found. Moreover, for the moving objects only about 50\% of their pixels had some matching points, and in some cases this number was as low as 19\%. This indicates how difficult the matching process is. 

Co-segmentation methods can also be condifered as a pre-processing for moving region detection. We 
applied one of the state-of-the-art co-segmentation methods (\cite{faktor2013co}) to our data set and show
that it  does not cope well with our data (see supplementary material for examples).

We also tested the applicability of back-projecting dense SFM for detecting moving regions. To this end, we use the SFM algorithm of \cite{wu2013visualsfm} on our data sets.
However, as can be seen in Fig.~\ref{fig:sfm}, the SFM algorithm reconstructs only a small number of scene points, and in the other two datasets it failed completely.
 Hence, it is impossible to use the back-projecting dense SFM to infer the dynamic regions unless all pixels that were not reconstructed are considered dynamic regions, which is clearly not the case. Further evidence that SFM methods struggle with our data sets is the failure of the SFM-based method of Wang {\em et al.} \cite{Wang2015} \footnote{Personal communication with the authors.}.

We could not compare our method to motion segmentation algorithms, because they work on video while we only use a sparse set of still images. Similarly, we could not compare our method to change detection algorithms, because our images were not taken from the same viewpoint, nor can they be aligned with a homography.

\section {Conclusions}
\label {sec:summary}
CrowdCam images are an emerging form of photography. Detecting moving regions is a basic step towards analyzing the dynamic content of such data. We proposed an algorithm that computes the probability of a pixel to be a projection of a dynamic 3D point. It does so by finding the probability that an epipolar patch (defined by a pair of matching epipolar lines) has matches  consistent with the epipolar geometry. This renders, our algorithm  less sensitive to matching errors than alternative algorithms that require precise matching. The aggregation of the results from a set of support images allows us to distinguish  dynamic regions from occluded regions and objects which move along epipolar lines. We evaluated our method on a new and challenging data set (that will be made public) and report results better than the alternative.

{\small
\bibliographystyle{ieee}
\bibliography{cite}
}

\end{document}